# Fake News Detection Using Majority Voting Technique

Dharmaraj R. Patil

Department of Computer Engineering, R.C. Patel Institute of Technology, Shirpur, Maharashtra, India

**Abstract**

Due to the evolution of the Web and social network platforms it becomes very easy to disseminate the information. Peoples are creating and sharing more information than ever before, which may be misleading, misinformation or fake information. Fake news detection is a crucial and challenging task due to the unstructured nature of the available information. In the recent years, researchers have provided significant solutions to tackle with the problem of fake news detection, but due to its nature there are still many open issues. In this paper, we have proposed majority voting approach to detect fake news articles. We have used different textual properties of fake and real news. We have used publicly available fake news dataset, comprising of 20,800 news articles among which 10,387 are real and 10,413 are fake news labeled as binary 0 and 1. For the evaluation of our approach, we have used commonly used machine learning classifiers like, Decision Tree, Logistic Regression, XGBoost, Random Forest, Extra Trees, AdaBoost, SVM, SGD and Naive Bayes. Using the aforementioned classifiers, we built a multi-model fake news detection system using Majority Voting technique to achieve the more accurate results. The experimental results show that, our proposed approach achieved accuracy of 96.38%, precision of 96%, recall of 96% and F1-measure of 96%. The evaluation confirms that, Majority Voting technique achieved more acceptable results as compare to individual learning technique.

## 1. Introduction

Fake news is information that is incorrect or misleading and presented as news. Fake news is not a new notion, but it is a prevalent occurrence in today's world. To identify false news, several ways have been proposed, including a linguistic approach, a topic-agnostic approach, a machine learning approach, a knowledge-based approach, and a hybrid approach [1]. The four viewpoints of fake news identification presented

by Zhou, X. et al. include the erroneous knowledge it carries, its writing style, its distribution patterns, and the legitimacy of its source [2]. On three separate datasets, Khan, J. Y., et al. conducted a benchmark study to examine the performance of several relevant machine learning algorithms [3]. The application of natural language processing algorithms for the identification of false news was investigated by R. Oshikawa et al. [4]. Molina, M. D. et al. have identified seven different types of online content under the label of "fake news" like false news, polarized content, satire, misreporting, commentary, persuasive information and citizen journalism [5]. Various surveys have been conducted by the researchers for the detection of false information, misinformation, misleading information and fake news and the detection methodlogies [6,7,8,9,10,11,12,13].

In this paper, we have proposed majority voting approach to detect fake news articles. We have used different textual properties of fake and real news. We have used publicly available fake news dataset, comprising of 20,800 news articles among which 10,387 are real and 10,413 are fake news labeled as binary 0 and 1. For the evaluation of our approach, we have uutilized commonly used machine learning classifiers like, Decision Tree (DT), Logistic Regression (LR), XGBoost (XGB), Random Forest (RF), Extra Trees (ET), AdaBoost (AB), Support Vector Machine (SVM), Stochastic Gradient Descent (SGD) and Naive Bayes (NB). Using the aforementioned classifiers, we built a multi-model fake news detection system using Majority Voting technique to achieve the more accurate results. The experimental results show that, our proposed approach achieved accuracy of 96.38%, precision of 96%, recall of 96% and F1-measure of 96%. The evaluation confirms that, Majority Voting technique achieved more acceptable results as compare to individual learning technique. The major contributions of this paper are as follow,

- We have proposed the majority voting based approach for the detection of fake news. It is found that, majority voting technique achieved more acceptable results as compare to individual learning technique.
- We have used various textual properties of fake and real news articles to classify the news more correctly as fake or real news.
- We have used Kaggle's publically available Fake News dataset to evaluate the performance of the state-of-the-arts machine learning classifiers like, Decision Tree, Logistic Regression, XGBoost, Random Forest, Extra Trees, AdaBoost,

- SVM, SGD and Naive Bayes. The classifier performance is evaluated using detection accuracy, precision, recall and F1-measure.
- We have built a multi-model learning system to detect the fake news and achieved more accurate results like, accuracy of 96.38%, precision of 96%, recall of 96% and F1-measure of 96%.

The remainder of this paper is organized as follows. Section 2 discusses brief related work on Fake News detection. Section 3 describes the proposed Methodology in detail. Section 4 discusses the experimental results and the performance of the proposed approach. Finally, in Section 5, the conclusion is presented.

## 2. Related Work

In recent various approaches have been proposed by the researchers towards fake news detection for Twitter, Facebook, Reddit, Wikipedia, YouTube and the newspapers. We described some of the approaches as follows.

Aldwairi, M. et al have utilized the features of social media title and the post to detect fake news. They have used logistic regression classifier and achieved 99.4% of accuracy [14]. Agarwal, V. et al. have used natural language processing and machine learning to solve the problem of fake news detection. They have used bag-of-words, n-grams, count vectorizer and TF-IDF to train the classifiers [15]. Nasir, J. A et al. have used hybrid CNN-RNN based deep learning approach to detect fake news [16]. Chauhan, T. et al. have used a deep learning based approach to differentiate false news from the original ones. They have used a LSTM neural network and a gloVe word embedding for vector representation of textual words. They have achieved an accuracy of 99.88% [17]. Shu, K. et al. have used datamining techniques to detect fake news [18]. Zhang, J. et al. have proposed a novel automatic fake news detection model, called FAKEDETECTOR, it is based on a set of explicit and latent features extracted from the textual information. It builds a deep diffusive network model to learn the representations of news articles, creators and subjects [19]. Nakamura, K. et al. have presented Fakeddit, a novel multimodal dataset consisting of over 1 million samples from multiple categories of fake news [20].

Kumari, S et al. have presented BERT based classification model to predict the domain and classification of fake news. They have achieved a macro F1 score of 83.76% [21]. Dong, X. et al. have proposed a novel Human-in-the-loop Based Swarm Learning (HBSL), to integrate user feedback into the loop of learning and inference for recognizing fake news without violating user privacy [22]. Liu, Y. et al. have proposed a novel model for early detection of fake news on social media through classifying news propagation paths [23]. Mertoğlu, U. et al. have developed an automated fake news detection system based on Turkish digital news content [24]. Agudelo, G. E. R. et al. have used machine learning and natural language processing for the identification of false news in public data sets [25].

Horne, B. D. et al. have examined the impact of time on state-of-the-art news veracity classifiers [26]. Zhou, X. et al. have proposed a theory-driven model for fake news detection. This method investigates news content at various levels: lexicon-level, syntax-level, semantic-level, and discourse-level [27]. Hansrajh, A. et al. have proposed a blended machine learning ensemble model developed from logistic regression, support vector machine, linear discriminant analysis, stochastic gradient descent, and ridge regression, which is then used on a publicly available dataset to predict if a news report is true or not [28].

Aslam, N. et al. have proposed an ensemble-based deep learning model to classify news as fake or real using LIAR dataset [29]. Ahmad, I., Yousaf et al. have proposed machine learning ensemble approach for automated classification of news articles. This study explores different textual properties that can be used to distinguish fake contents from real [30]. Torabi Asr, F. et al. have introduced the MisInfoText repository to the community. They have performed a topic modelling experiment to elaborate on the gaps and sources of imbalance in currently available datasets to guide future efforts [31].

Islam, N. et al. have proposed a novel solution by detecting the authenticity of news through natural language processing techniques. Also proposed a novel scheme comprising three steps, namely, stance detection, author credibility verification, and machine learning-based classification, to verify the authenticity of news. They have achieved an accuracy of 93.15%, precision of 92.65%, recall of 95.71%, and F1-score of 94.15% for the support vector machine algorithm [32]. Lai, C. M. et al. have used the combination of ML and NLP to classify fake news based on an open, large and labeled corpus on Twitter. They have compared several state-of-the-art ML and neural network models based on content-only features [33]. Stissi, S has created a functional, accessible, human-centered user interface for automatic fake news detection [34]. Groh, M., Epstein et al. have compared the performance of ordinary human observers with the leading computer vision deepfake detection model and find them similarly accurate, while making different kinds of mistakes [35].

Thota, A. et al. have presented neural network architecture to accurately predict the stance between a given pair of headline and article body. They have archived an accuracy of 94.21% on test data [36]. Yun, T. U. et al. have proposed to detect Korean fake news using text mining and machine learning techniques [37]. Mazzeo, V. et al. have proposed to detect potential misleading and fake contents by capturing and analyzing textual information, which flow through search engines [38]. Gangireddy, S. C. R. et al. have used the task of unsupervised fake news detection, which considers fake news detection in the absence of labelled historical data [39]. Fung, Y. et al. have proposed a novel benchmark for fake news detection at the knowledge element level, as well as a solution for this task which incorporates cross-media consistency checking to detect the fine-grained knowledge elements making news articles misinformative [40].

Tschiatschek, S. et al. have considered leveraging crowd signals for detecting fake news and is motivated by tools recently introduced by Facebook that enable users to flag fake news [41]. Preston, S. et al. have proposed an approach to assess whether individuals who show high levels of 'emotional intelligence' are less likely to fall for fake news items [42]. Shu, K. et al. have developed a sentence-comment co-attention sub-network to exploit both news contents and user comments to jointly capture explainable top-k check-worthy sentences and user comments for fake news detection [43]. Meesad, P. has proposed a framework for robust Thai fake news detection. The framework comprises three main modules, including information retrieval, natural language processing, and machine learning [44]. Guibon, G. et al. have compared different automatic approaches for fake news detection based on statistical text analysis on the vaccination fake news dataset provided by the Storyzy company [45].

Nguyen, D. M. et al. have developed a graph-theoretic method that inherits the power of deep learning while at the same time utilizing the correlations among the articles. They have formulated fake news detection as an inference problem in a Markov random field (MRF) which can be solved by the iterative mean-field algorithm [46]. Pandey, S. et al. have used algorithms such as K-Nearest Neighbor, Support Vector Machine, Decision Tree, Naïve Bayes and Logistic Regression Classifiers to identify the fake news from real ones in a given dataset and also have increased the efficiency of these algorithms by pre-processing the data to handle the imbalanced data more appropriately [47]. Nordberg, P. et al. have identified key methods for automatic fake news detection in order to lay the foundation for end-user support system designed to help users identify and avoid fake news [48]. Ashraf, N.et al. have reported several machine learning classifiers on the CLEF2021 dataset for the tasks of news claim and topic classification using n-grams [49]. Pérez-Rosas, V. et al. have focused on the automatic identification of fake content in online news. They have introduced two novel datasets for the task of fake news detection, covering seven different news domains [50].

## 3. Methodology

### 3.1. Proposed Fake News Detection System

Fig. 1. shows our proposed Fake News detection system. It has stages like pre-processing, feature extraction and learning.

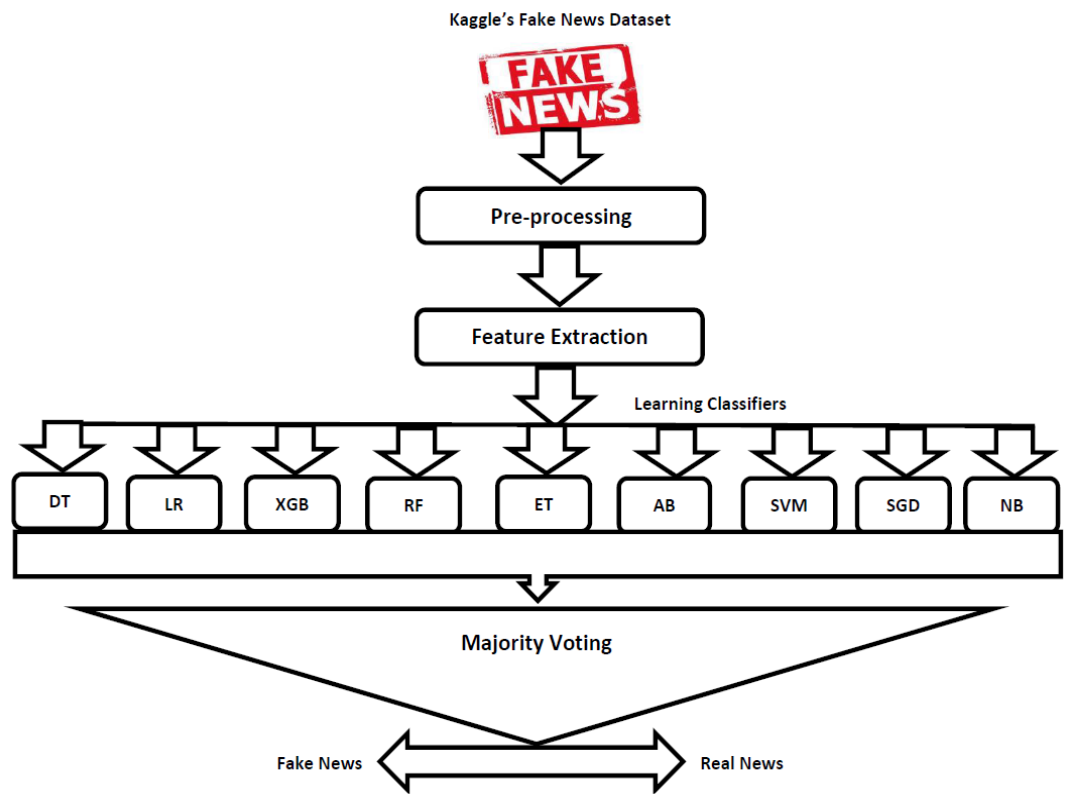

Fig.1. Proposed Framework of our Fake News Detection System

### 3.2. Pre-processing

We passed the raw Fake News dataset from Kaggle to the Python pre-processing module. We employ a text preparation technique that includes:

- Punctuation Removal,
- Tokenization,
- Stop Word Removal,
- Word Stemming,
- URL Removal, and
- Names to eliminate undesired items from the dataset.

### 3.3. Feature Extraction

The processed text is then passed on to feature extraction; which extracts features such as TF-IDF weights and CountVectorizer.

### 3.3.1. TF_IDF Features

The TF_IDF statistic is intended to assess the relevance of a word in a set of texts (or corpus). It is represented by an equation (1). The frequency of the term, TF (t, d), is the frequency of occurrence of the term t internal document d.

$$TF(t,d) = \frac{f_{t,d}}{\sum_{t' \in d} f_{t',d}} \quad (1)$$

where,
$f_{t,d}$ is the raw count of a term in a document, i.e., the number of times that term t occurs in document d.

A word's inverse document frequency represents the fraction of documents in the corpus that include the term. As seen in the equation (2), words that are unique to a small percentage of documents have greater relevance values than terms that are common to all documents,

$$IDF(t,D) = \log \frac{N}{|\{d \in D : t \in d\}|} \quad (2)$$

where,
N: total number of documents in the corpus.
$|\{d \in D : t \in d\}|$ : number of documents where the term $t$ appears.
The Term frequency–inverse document frequency (TF-IDF) is calculated using equation (3),

$$TFIDF(t,d,D) = TF(t,d) \cdot IDF(t,D) \quad (3)$$

### 3.3.2. CountVectorizer Features

CountVectorizer tokenizes (tokenization means breaking down a sentence or paragraph or any text into words) the text along with performing very basic preprocessing like removing the punctuation marks, converting all the words to lowercase, etc. The vocabulary of known words is formed which is also used for encoding unseen text later. An encoded vector is returned with a length of the entire vocabulary and an integer count for the number of times each word appeared in the document.

**3.4. Machine Learning Classifiers for Fake News Detection**

We have evaluated the performance of 9 machine learning classifiers for the detection of Fake News, including Decision Tree (DT), Logistic Regression (LR), XGBoost (XGB), Random Forest (RF), Extra Trees (ET), AdaBoost (AB), Support Vector Machine (SVM), Stochastic Gradient Descent (SGD) and Naive Bayes (NB). From these 9 machine learning classifiers we have built a multi-model classifier using majority voting for the final decision. We have used the scikit-learn python library implementation of each classifier [51]. The brief discussion of each classifier is given as below.

*3.4.1. Decision Tree*

One of the most popular classification approaches is decision tree learning. It is highly efficient and has classification accuracy comparable to other learning methods. A decision tree is a tree that reflects the classification model that has been learned. It's an easy-to-understand decision tree classification paradigm. The method evaluates all feasible data split tests and chooses the one with the highest information gain [52,53].

*3.4.2. Logistic Regression*

Logistic regression is a popular Machine Learning algorithm that is used in the Supervised Learning approach. It is used to predict the categorical dependent variable from a set of independent variables. The outcome of a categorical dependent variable is predicted using logistic regression. As a result, the outcome must be a categorical or discrete value. It can be Yes or No, 0 or 1, true or False, and so on, but instead of displaying precise values like 0 and 1, it offers probability values that lie between 0 and 1 [54,55].

*3.4.3. XGBoost*

The XGBoost (eXtreme Gradient Boosting) approach is well-known and successful. Gradient boosting is a supervised learning approach that combines estimates from a series of simpler and weaker models to properly predict a target variable. Because of its strong handling of a wide range of data kinds, relationships, and distributions, as well as the huge range of hyperparameters that can be fine-tuned, the XGBoost technique performs well in machine learning issues. XGBoost is capable of dealing with regression, classification (binary and multiclass), and ranking issues [56].

*3.4.4. Random Forest*

A random forest, as the name implies, is composed of a huge number of individual decision trees that collaborate as an ensemble. The random forest generates a class prediction for each tree, and the class with the highest votes becomes the forecast of our model. The Random Forest's basic principle is communal knowledge, which is both simple and powerful. The random forest model is particularly successful because it is composed of a large number of largely uncorrelated models (trees) that collaborate to outperform each of the individual constituent models [57,58].

*3.4.5. Extra Tree*

Extra trees classifier is type of ensemble learning techniques that aggregates the classification results of several non-correlated decision trees gathered in "Forest" to obtain its classification results. It is conceptually very similar to Random Forest Classifier and varies mostly in the manner the decision trees in the forest are formed. The Extra Trees Forest's Decision Trees are constructed from training samples. Then, at each test node, each tree is given random samples of k feature from the feature sets. From which each of the decision tree must select the best features to divide data using important mathematical criterion. This random selection of features results in the construction of several de-correlated decision trees [59,60].

*3.4.6. AdaBoost*

AdaBoost is the most frequently used and researched algorithm, with applications in a wide range of fields. Freund and Schapire created the AdaBoost algorithm in 1995. Abstract Boosting is a machine learning strategy that combines a large number of weak and incorrect classifiers to generate a highly accurate classifier. It's simple to use, quick, and simple to comprehend. It does not require any previous information from the weak learner, hence it may be utilised in combination with any weak hypothesis identification technique [61,62].

*3.4.7. Support Vector Machine*

The Support Vector Machine (SVM) is a supervised machine learning technology that may be used to handle classification and regression issues. It is, however, mostly used in classification difficulties. Each data item is represented as a point in n-dimensional space, with the value of each feature being the value of a specific coordinate in the SVM algorithm. Then, we achieve classification by selecting the hyper-plane that best separates the two classes. Individual observation coordinates are utilised to compute support vectors. The SVM classifier is a frontier that best differentiates between the two classes (hyper-plane/line) [63,64].

*3.4.8. Stochastic Gradient Descent*

Stochastic Gradient Descent (SGD) is a simple yet very efficient approach to fitting linear classifiers and regressors under convex loss functions such as (linear) Support Vector Machines and Logistic Regression. Even though SGD has been around in the machine learning community for a long time, it has received a considerable amount of attention just recently in the context of large-scale learning. SGD has been successfully applied to large-scale and sparse machine learning problems often encountered in text classification and natural language processing. Stochastic Gradient Descent (SGD) classifier basically implements a plain SGD learning routine supporting various loss functions and penalties for classification. Scikit-learn provides SGDClassifier module to implement SGD classification [65].

*3.4.9. Naïve Bayes*

A Naive Bayes classifier is a probabilistic machine learning model that's used for classification task. The crux of the classifier is based on the Bayes theorem.

$$P(A|B) = \frac{P(B|A)P(A)}{P(B)} \tag{4}$$

Using Bayes theorem, we can find the probability of A happening, given that B has occurred. Here, B is the evidence and A is the hypothesis. The assumption made here is that the predictors/features are independent. That is presence of one particular feature does not affect the other. Hence it is called naïve [66].

*3.4.10. Majority Voting*

Voting is the most basic ensemble strategy, and it is usually quite effective. It may be applied to classification and regression issues. In this scenario, it divides a model into two or more sub-models, in this case five. To integrate predictions from each sub-model, the majority voting technique is employed. Figure 3 depicts the majority voting process. It is a meta-classifier that identifies machine learning classifiers that are conceptually similar or different using a majority vote. We anticipate the final class label using majority voting, which is the class label that classification algorithms most commonly predict. Using equation (4) and the majority vote of each classifier $C_j$, we predict the class label y [67, 68].

$$y = mode\{C1(x), C2(x), ..., Cm(x)\} \tag{5}$$

where,
y = predicted class label and
C1(x), C2(x),..., Cm(x)=classification models.

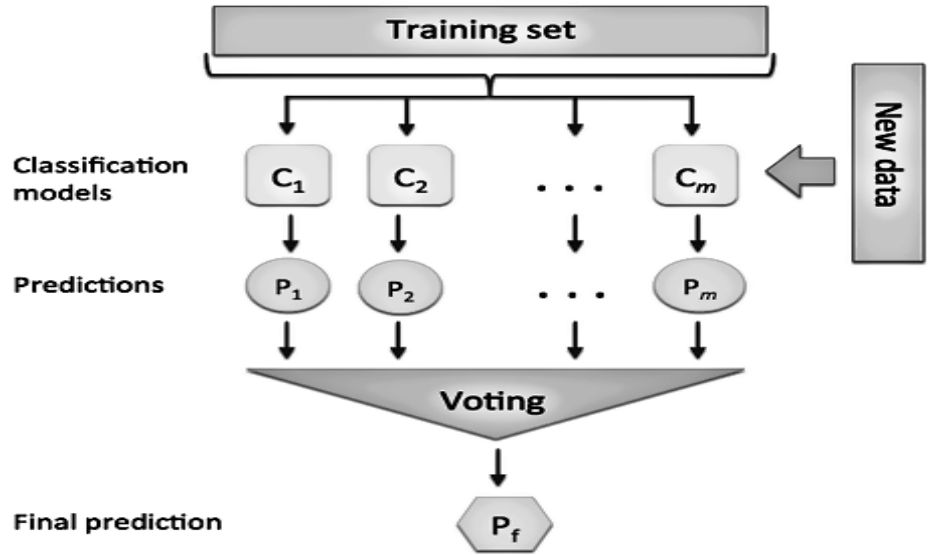

Fig.2. Majority Voting Classifier

## 4. Experimental Setup and Evaluation

### 4.1. Dataset and Data Source

We utilized the Fake News Kaggle dataset, which comprising of 20,800 news articles among which 10,387 are real and 10,413 are fake news labeled as binary 0 and 1 [69]. For training and testing, the classifiers divided the dataset in an 80:20 ratios.

### 4.2. Evaluation Measures

We utilized the following metrics to assess classifier performance. A binary classifier labels all data elements in a test dataset with a 0 or 1. True positive ($TP_F$), true negative ($TN_F$), false positive ($FP_F$), and false negative ($FN_F$) are the four results of this classification [70]. The following equations are used to compute the $Accuracy_F$, $Precision_F$, $Recall_F$, and $F1-score_F$ measures.

$$Accuracy_F = \frac{TP_F + TN_F}{TP_F + TN_F + FP_F + FN_F} \quad (6)$$

$$Precision_F = \frac{TP_F}{TP_F + FP_F} \quad (7)$$

$$Recall_F = \frac{TP_F}{TP_F + FN_F} \quad (8)$$

$$F1-score_F = 2 \cdot \frac{Precision_F \cdot Recall_F}{Precision_F + Recall_F} \quad (9)$$

*4.3. Performance Evaluation of classifiers on Fake News dataset*

Table 1 and Fig. 3 shows the performance of the classifiers on the Kaggle Fake News dataset in terms of accuracy, precision, recall and F1-score. The Decision Tree classifier has achieved the accuracy of 88.51%, precision of 89%, recall of 89% and F1-score of 89%. The Logistic Regression classifier achieved the accuracy of 94.89%, precision of 95%, recall of 95% and F1-score of 95%. The XGBoost classifier achieved the accuracy of 95.79%, precision of 96%, recall of 96% and F1-score of 96%. The Random Forest classifier achieved the accuracy of 91.60%, precision of 92%, recall of 92% and F1-score of 92%. The AdaBoost classifier achieved the accuracy of 92.72%, precision of 93%, recall of 93% and F1-score of 93%. The Support Vector Machine classifier achieved the accuracy of 96.49%, precision of 97%, recall of 96% and F1-score of 96%. The Extra Trees classifier achieved the accuracy of 91.11%, precision of 92%, recall of 91% and F1-score of 91%. The SGD classifier achieved the accuracy of 96.11%, precision of 96%, recall of 96% and F1-score of 96%. The Naïve Bayes classifier achieved the accuracy of 84.62%, precision of 88%, recall of 85% and F1-score of 84%. Our proposed Multi-model classifier using the majority voting achieved the accuracy of 96.38%, precision of 96%, recall of 96% and F1-score of 96%. It is found that, our proposed approach achieved more stable and accurate results in comparison with aforementioned classifiers.

Table 1. Performance Evaluation of classifiers on Fake News dataset

| Classifier | Accuracy (%) | Precision (%) | Recall (%) | F1-score (%) |
|---|---|---|---|---|
| DT | 88.51 | 89 | 89 | 89 |
| LR | 94.89 | 95 | 95 | 95 |
| XGBoost | 95.79 | 96 | 96 | 96 |
| RF | 91.6 | 92 | 92 | 92 |
| AdaBoost | 92.72 | 93 | 93 | 93 |
| SVM | 96.49 | 97 | 96 | 96 |
| ET | 91.11 | 92 | 91 | 91 |
| SGD | 96.11 | 96 | 96 | 96 |
| NB | 84.62 | 88 | 85 | 84 |
| **Proposed Approach** | **96.38** | **96** | **96** | **96** |

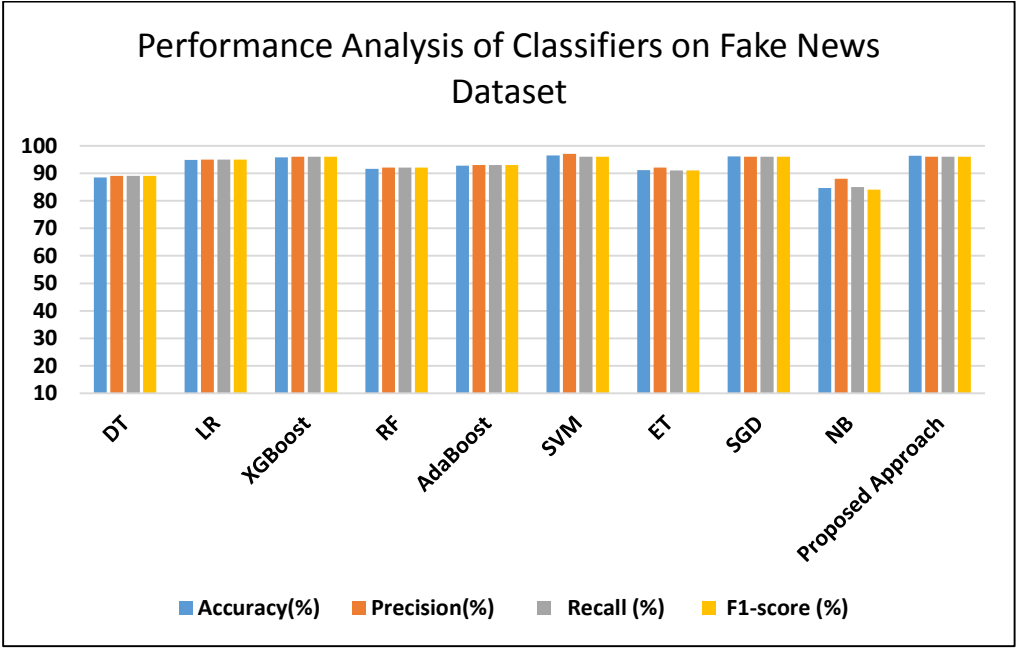

Fig.3. Performance Analysis of Classifiers on Fake News Dataset

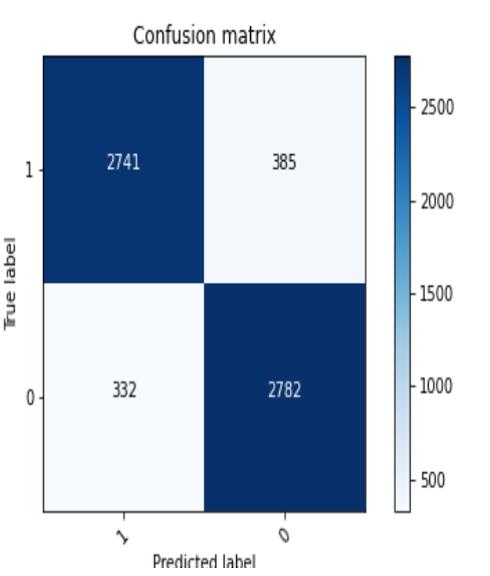
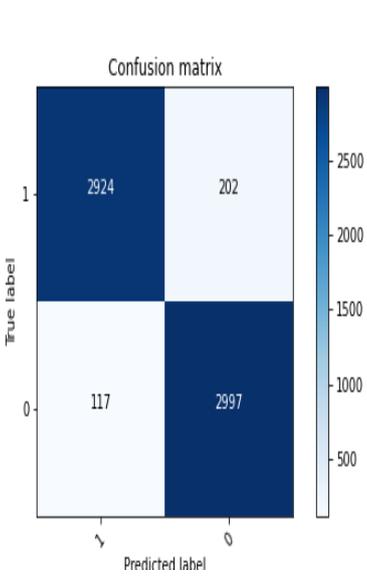

(a)                                        (b)

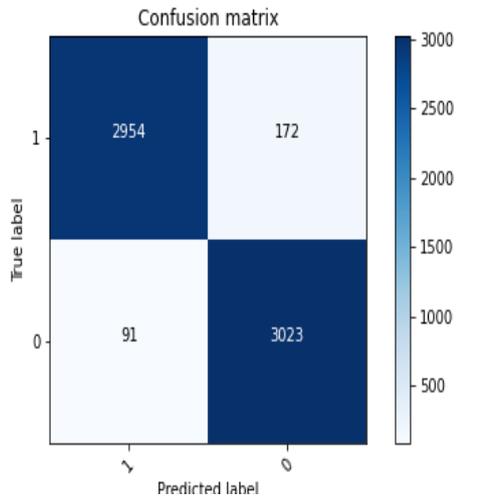

(c)

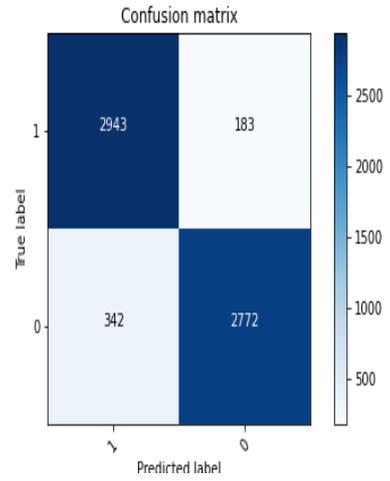

(d)

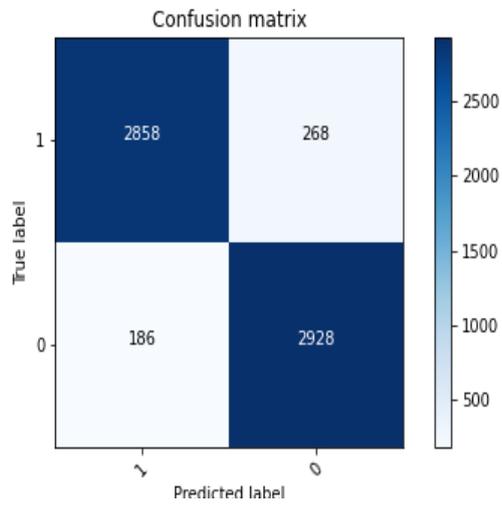

(e)

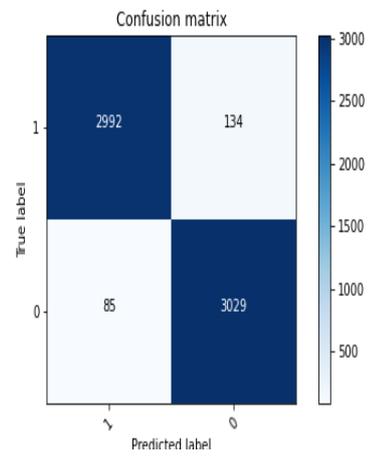

(f)

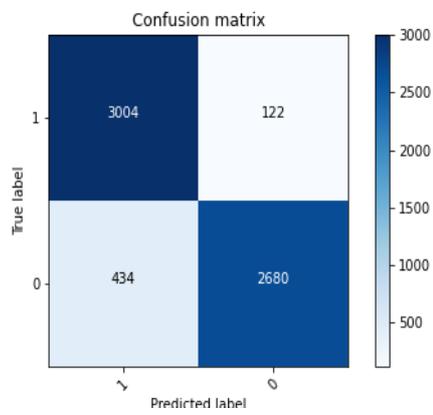

(g)

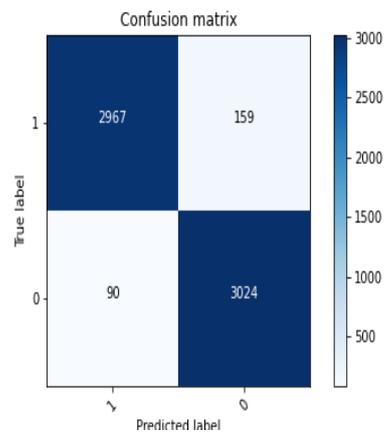

(h)

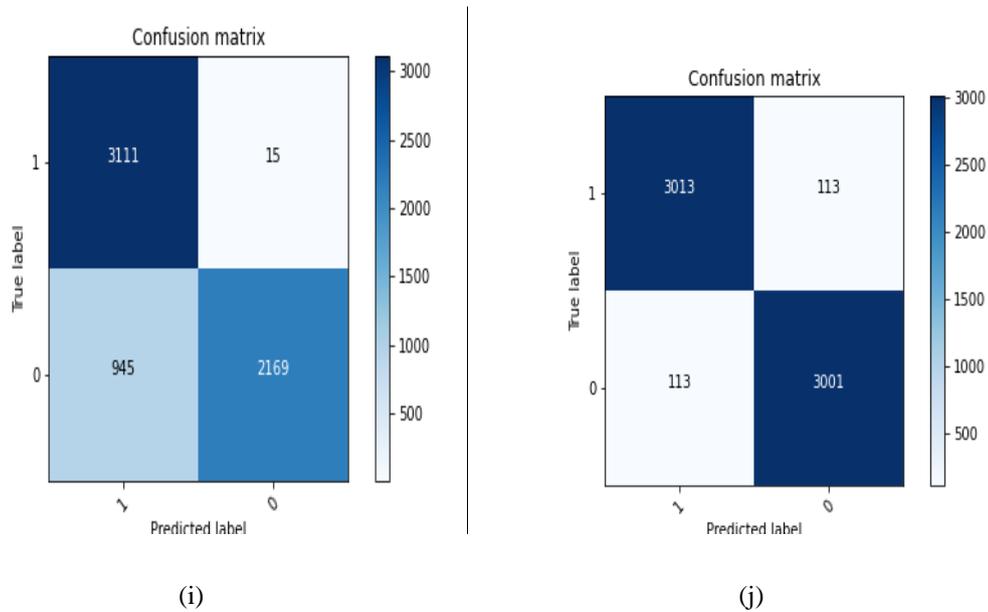

(i)  (j)

Fig. 4. Confusion Matrix for (a) Decision Tree (DT), (b) Logistic Regression (LR), (c) XGBoost (XGB), (d) Random Forest (RF), (e) AdaBoost (AB), (f) Support Vector Machine (SVM), (g) Extra Trees (ET), (h) Stochastic Gradient Descent (SGD), (i) Naive Bayes (NB) and (j) proposed approach using majority voting.

Fig.4. shows the confusion matrix for the classifiers including, Decision Tree (DT), Logistic Regression (LR), XGBoost (XGB), Random Forest (RF), AdaBoost (AB), Support Vector Machine (SVM), Extra Trees (ET), Stochastic Gradient Descent (SGD), Naive Bayes (NB) and proposed approach using majority voting. Fig.5. shows the ROC curve for the classifiers.

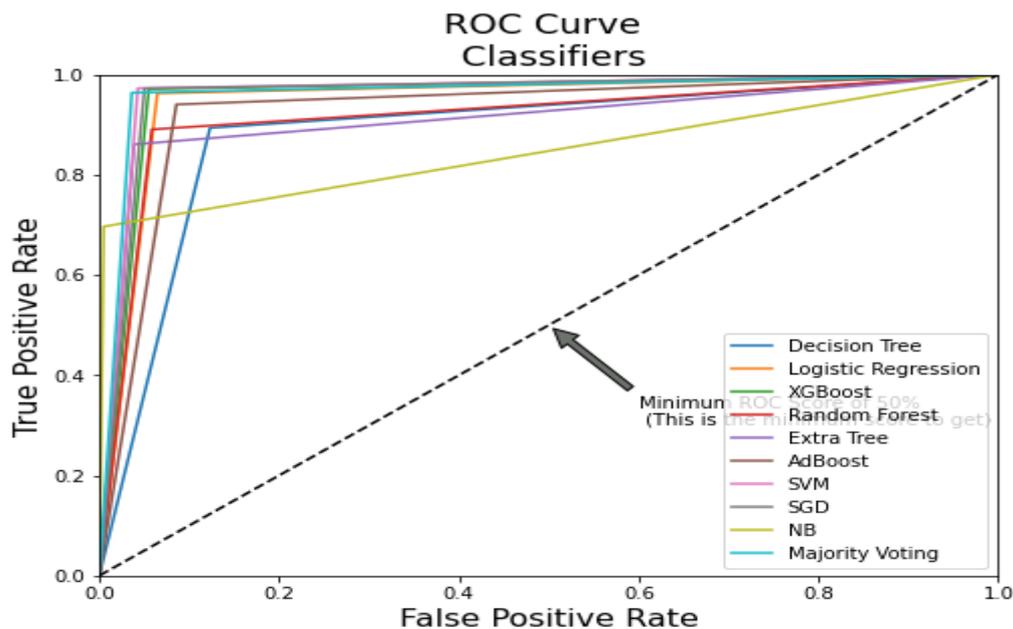

Fig.5. ROC Curve of the Classifiers on Fake News Dataset

## 5. Conclusions

In this paper, we have proposed a majority voting based approach for the detection of Fake News. We have used publically available Kaggle's Fake News dataset, which comprising of 20,800 news articles among which 10,387 are real and 10,413 are fake news labeled as binary 0 and 1. We have utilized commonly used machine learning classifiers such as Decision Tree, Logistic Regression, XGBoost, Random Forest, AdaBoost, Support Vector Machine, Extra Trees, Stochastic Gradient Descent and Naive Bayes to analyze the dataset. We have used TF-IDF and CountVectorizer Features. Using the aforementioned classifiers and features, we built a multi-model fake news detection system using Majority Voting technique to achieve the more accurate results. The experimental results show that, our proposed approach achieved accuracy of 96.38%, precision of 96%, recall of 96% and F1-measure of 96%. The evaluation confirms that, Majority Voting technique achieved more accurate results as compare to individual learning technique.